\documentclass{article}
\usepackage{ijcai24}

\usepackage{times}
\usepackage{soul}
\usepackage{color}
\usepackage{url}
\usepackage{amsfonts,amssymb}
\usepackage[hidelinks]{hyperref}
\usepackage[utf8]{inputenc}
\usepackage[small]{caption}
\usepackage{graphicx}
\usepackage{amsmath}
\usepackage{amsthm}
\usepackage{booktabs}
\usepackage{algorithm}
\usepackage{xcolor} 
\usepackage{algorithmic}
\usepackage[switch]{lineno}
\usepackage{booktabs}
 \usepackage{multirow}
 \usepackage{pifont}
 \usepackage{bm}
 \usepackage{tabularx}
 \usepackage{adjustbox}
 \usepackage{makecell}
\usepackage{array}  

\usepackage{multicol}

\title{A Survey on Visual Anomaly Detection: Challenge, Approach, and Prospect}

\author{
Yunkang Cao$^{1}$\thanks{Equal contribution.}
\and
Xiaohao Xu$^{2}$\footnotemark[1]
\and
Jiangning Zhang$^{3}$\footnotemark[1]
\and
Yuqi Cheng$^1$
\and
\\Xiaonan Huang$^2$
\and
Guansong Pang$^4$
\and
Weiming Shen$^{1}$\thanks{Corresponding author.}
\affiliations
$^1$Huazhong University of Science and Technology,
$^2$University of Michigan, Ann Arbor\\
$^3$Youtu Lab, Tencent,
$^4$Singapore Management University\\
\emails
\{cyk\_hust, yuqicheng, shenwm\}@hust.edu.cn,
\{xiaohaox, xiaonanh\}@umich.edu,\\
186368@zju.edu.cn,
gspang@smu.edu.sg
}

\begin{document}

\maketitle

\begin{abstract}
Visual Anomaly Detection (VAD) endeavors to pinpoint deviations from the concept of normality in visual data, widely applied across diverse domains, \textit{e.g.}, industrial defect inspection, and medical lesion detection.
This survey comprehensively examines recent advancements in VAD by identifying three primary challenges: 1) scarcity of training data, 2) diversity of visual modalities, and 3) complexity of hierarchical anomalies. 
Starting with a brief overview of the VAD background and its generic concept definitions, we progressively categorize, emphasize, and discuss the latest VAD progress from the perspective of sample number, data modality, and anomaly hierarchy.  
Through an in-depth analysis of the VAD field, we finally summarize future developments for VAD and conclude the key findings and contributions of this survey.

\end{abstract}
\section{Introduction}
Visual anomaly detection (VAD) stands as a pivotal task spanning diverse domains~\cite{pang_deep_2022}, involving the identification of deviations in visual data from established normality. 
In recent years, we have seen notable progress in this field across multiple domains. 
For instance, inspecting defects in industrial settings~\cite{MVTec-AD}, identifying lesions in medical image analysis~\cite{antonelli2022medical,2023liuST}, and detecting unknown objects in autonomous driving scenarios~\cite{bogdoll2022anomaly}. The significance of VAD extends beyond these specific applications, as its ability to uncover irregularities in visual data contributes significantly to enhancing the overall reliability and safety of various technological systems.
Despite great progress, VAD still encounters three predominant challenges as illustrated in Figure~\ref{fig:teaser}: 

\begin{figure}[t!]
\centering\includegraphics[width=1.\linewidth]{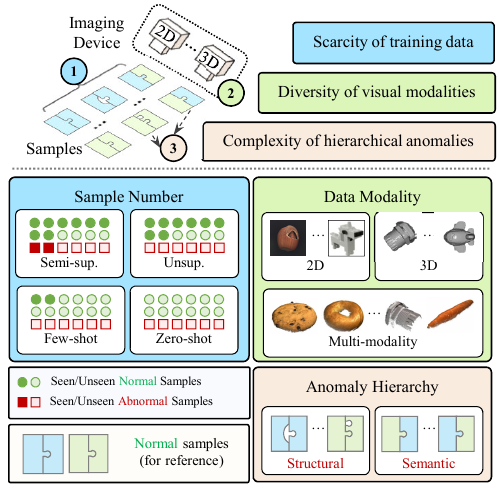}\vspace{-3mm}
\caption{Major VAD challenges (\textbf{Top}) and taxonomies (\textbf{Bottom}).}
\vspace{-3mm}
\label{fig:teaser}
\end{figure}




\noindent\textbf{1) Scarcity of Training Data.} 
Practical VAD systems often struggle to amass abundant abnormal samples for training~\cite{MVTec-AD}. 
In special application scenarios, normal samples may even be inaccessible due to data privacy concerns~\cite{WinClip}. 
The scarcity of training data poses a significant challenge in VAD, requiring the training of models from partially observed samples in order to subsequently detect anomalies in an open-world environment.

\noindent\textbf{2) Diversity of Visual Modalities.} 
VAD systems employ diverse imaging devices to capture visual information, such as color cameras~\cite{MVTec-AD} and radar scanners~\cite{MVTec-3D}. 
The utilization of these diverse imaging techniques introduces distinct visual modalities, presenting both utility and complexity in their effective integration.


\noindent\textbf{3) Complexity of Hierarchical Anomalies.} 
Anomalies may manifest in various hierarchies.
Some structural anomalies (\textit{e.g.}, visual scratch) can be identified through local regions, while others semantic anomalies (\textit{e.g.}, logical mismatch) demand a higher-level understanding of the normal context~\cite{MVTec-LOCO}. 
It is challenging for VAD models to concurrently possess both fine-grained and global understanding of visual data.


The above challenges continually drive the research frontiers of the VAD field. 
In this survey, we aim to comprehensively analyze the latest progress made in tackling these challenges. 
After providing a concise overview of the background of VAD (Sec.~\ref{sec:2}), we will review previous achievements and research trends from perspectives corresponding to the aforementioned challenges: 1) sample number (Sec.~\ref{sec:3.1}), 2) data modality (Sec.~\ref{sec:3.2}), and 3) anomaly hierarchy (Sec.~\ref{sec:3.3}). 
Afterward, we will present some potential and emerging future research directions in Sec.~\ref{sec:4}. 
Finally, Sec.~\ref{sec:5} will conclude this survey by summarizing key findings and contributions.

\section{Background}\label{sec:2}


In this section, we briefly review the background of VAD, encompassing the conceptual definitions and a generic formulation for VAD. Then we outline prominent datasets and metrics used for evaluating VAD methods and finally introduce relevant surveys to clarify the contributions of this survey.

\noindent\paragraph{Concept Definition.} 
Here, we formally define concepts of visual data, anomalies, and the VAD task. 


\textbf{1) Visual data} is classified into four fundamental categories: \textit{data point}, \textit{entity}, \textit{relation}, and \textit{frame}. i) A data point $D$ represents the smallest discernible element captured by imaging devices, such as a pixel in an image or a point in a point cloud. ii) An entity $E$ is a cohesive set of data points that collectively represent a real-world object. Denoted as $E = \{D_1, D_2, \ldots, D_n\}$, an entity encompasses individual data points that together form a meaningful visual element. iii) A relation function $\varphi$ takes multiple entities $\{E_1, E_2, \ldots, E_m\}$ as input and combines them to form a visual frame. iv) A frame $F = \varphi(E_1, E_2, \ldots, E_m)$ encapsulates the interrelation between different entities, capturing contextual information within the visual scene.

\textbf{2) Anomaly concept} refers to observations that deviate from the concept of normality~\cite{ruff_unifying_2021}. Anomalies in visual data exhibit a hierarchical relationship, where anomalies at lower levels can propagate to higher levels. For instance, errors in individual data points $D$ lead to anomalies in the formation of entities $E$. This hierarchical structure enables a nuanced understanding of anomalies across different granularities within visual data. Considering the hierarchies of anomalies, we define two types of anomalies: 
\textbf{\textit{i) Structural anomalies}} focus on the integrity of individual data points and their organization within entities, as exemplified in MVTec AD~\cite{MVTec-AD}. These anomalies are particularly insightful for detecting local structure deviations, such as lesions in medical visual data or defects in industrial inspections. 
\textbf{\textit{ii) Semantic anomalies}}, on the other hand, encompass deviations at higher hierarchical levels, including the entity, relation, and frame levels. Anomalies at the entity level may involve the misinterpretation of entities within the visual scene, such as unknown objects on the road~\cite{bogdoll2022anomaly}. Relation-level anomalies involve inaccuracies in the contextual connections between entities, as illustrated in MVTec LOCO~\cite{MVTec-LOCO}. Frame-level anomalies denote abnormalities in the overall visual acquisition, typically presented as novelty detection~\cite{ruff_unifying_2021} or one-class classification~\cite{IGD}. 
Addressing semantic anomalies is vital for understanding the context of visual entities and their interrelations, ensuring accurate and meaningful interpretations of the entire scene.
\begin{figure}[t]
\centering\includegraphics[width=\linewidth]{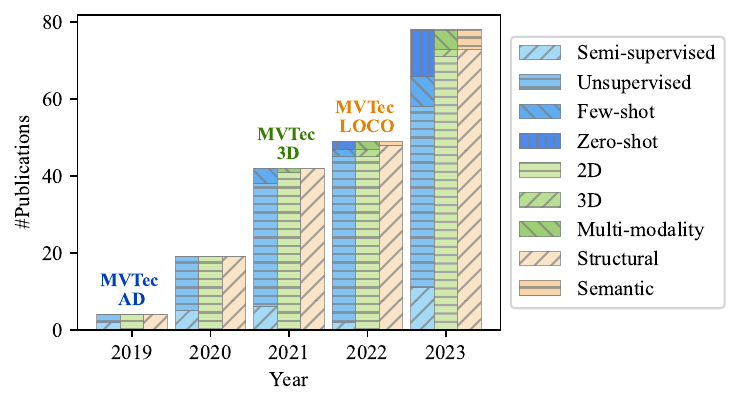}\vspace{-3mm}
\caption{
Number of VAD publications regarding taxonomies under the three perspectives in Sec.~\ref{sec:3}. \textcolor[RGB]{0,63,188}{Blue} for Sec.~\ref{sec:3.1}, \textcolor[RGB]{56,125,0}{green} for Sec.~\ref{sec:3.2}, and \textcolor[RGB]{230,129,0}{orange} for Sec.~\ref{sec:3.3}.
}
\label{fig:work_number}
\end{figure} 

\begin{table}[t]
    \centering
    \caption{Comparison with representative VAD surveys. \ding{52}: Included. $\bm{+}$: Partially Included. \ding{55}: Not Included.}
    \label{tab:survey}
    \renewcommand{\arraystretch}{1.1}
    \setlength\tabcolsep{3.0pt}
    \vspace{-2mm}
    \resizebox{1.0\linewidth}{!}{
        \begin{tabular}{p{3.8cm}>{\raggedright} p{1.6cm}<{\centering} p{1.6cm}<{\centering} p{1.6cm}<{\centering}}
            \toprule[1.5pt]
            \multirow{2}{*}{Survey}     & \multirow{2}{*}{\makecell[c]{Sample \\Number}} & \multirow{2}{*}{\makecell[c]{Data \\Modality}} & \multirow{2}{*}{\makecell[c]{Anomaly \\Hierarchy}} \\
            & & & \\
            \hline
            \cite{ruff_unifying_2021}   & $\bm{+}$      & \ding{55}     & \ding{55} \\
            \cite{pang_deep_2022}       & $\bm{+}$      & \ding{55}     & \ding{55} \\
            \cite{tao_2022}             & $\bm{+}$      & \ding{55}     & \ding{55} \\
            \cite{diers_survey_2023}    & $\bm{+}$      & \ding{55}     & \ding{55} \\
            \cite{liu2023deep}          & $\bm{+}$      & $\bm{+}$      & \ding{55} \\
            \hline
            Ours                        & \ding{52}    & \ding{52}    & \ding{52} \\
            \bottomrule[1.5pt]
        \end{tabular}
    }
    \vspace{-5mm}
\end{table}

\textbf{3) Visual anomaly detection} aims to develop a model for detecting visual anomalies. The task involves a training dataset $\mathcal{F}_{\text{train}}$ comprising both normal frames ${F}_{{n}}$ and abnormal frames ${F}_{{a}}$, each accompanied by corresponding ground truths expressed as hard labels. The principal goal is to establish a discriminative function $f_\theta: {F} \to [0,1]$, parameterized by $\theta$, using $\mathcal{F}_{\text{train}}$ to precisely assign anomaly scores to unlabeled frames in $\mathcal{F}_{\text{test}}$. Recent advancements highlight the necessity for more detailed anomaly scores, extending to data point-level, entity-level, and relation-level assessments.

\noindent\paragraph{Scope of This Survey.} Notably, VAD has witnessed substantial advancements in the industrial domain compared to other domains. Despite methodological variations across domains, the fundamental principles governing VAD exhibit considerable consistency. Consequently, this survey strategically concentrates on VAD within industrial scenarios as a representative, endeavoring to furnish a meticulous and exhaustive review of the entire VAD landscape. Notably, the semantic anomalies within industrial scenarios predominantly manifest at the relation level, posing challenges and garnering escalating attention.

\noindent\paragraph{Datasets \& Metrics.} 
Recent advancements in VAD are significantly influenced by several datasets, such as MVTec AD~\cite{MVTec-AD}, MVTec 3D~\cite{MVTec-3D}, MVTec LOCO~\cite{MVTec-LOCO}, and VisA~\cite{VisA}. 
The alignment between predicted probability distributions and true probability distributions provided by these datasets can evaluate the performance of VAD methods. Various metrics~\cite{MVTec-AD,WinClip} are employed for this purpose, encompassing Area Under the Receiver Operating Characteristic curve (AUROC) and Area Under the Per-Region-Overlap curve (AUPRO), among others.


\begin{table*}[ht]
\centering
\caption{Summary of existing AD methods from three perspectives, addressing the aforementioned challenges. For each sub-setting, we list several representatives. \ding{52}: Involved. $\bm{+}$: Partially Involved. \ding{55}: Not Involved. \textcircled{ }: Disregarded. }
\label{tab:summary}
\vspace{-2mm}
\resizebox{\linewidth}{!}{
\begin{tabular}{@{}clccccccll@{}}
\toprule[1.5pt]
\multirow{2}{*}{Perspective}   & \multirow{2}{*}{Taxonomy} & \multicolumn{2}{c}{Training Set}   & \multicolumn{2}{c}{Modality} & \multicolumn{2}{c}{Hierarchy}    & \multicolumn{1}{c}{\multirow{2}{*}{Representative Works}} & \multicolumn{1}{c}{\multirow{2}{*}{Summary}} \\ \cmidrule(lr){3-4} \cmidrule(lr){5-6} \cmidrule(lr){7-8}
 &  & \multicolumn{1}{c}{${F}_n$} & \multicolumn{1}{c}{${F}_a$} & \multicolumn{1}{c}{RGB} & \multicolumn{1}{c}{3D} & \multicolumn{1}{c}{Structural} & \multicolumn{1}{c}{Semantic} & \multicolumn{1}{c}{}  & \multicolumn{1}{c}{}\\ \midrule
\multirow{4}{*}{\begin{tabular}[c]{@{}c@{}}\makecell[c]{Sec.~\ref{sec:3.1} \\ Sample Number} \end{tabular}}   & 
Semi-supervised &   \ding{52}  &$\bm{+}$    &    \textcircled{ }    & \textcircled{ }   & \textcircled{ } &  \textcircled{ }   &   DRA, PRN, BiaS, BGAD & Preventing Overfitting   \\
 & Unsupervised&   \ding{52}  &\ding{55}   &    \textcircled{ }    & \textcircled{ }   &\textcircled{ } &  \textcircled{ }  &   ST, Draem, PatchCore, RD4AD & Modelling feature distribution\\
 & Few-shot    &   $\bm{+}$   &\ding{55}  &    \textcircled{ }    & \textcircled{ }   & \textcircled{ } &  \textcircled{ }  &  RegAD, GraphCore, FastRecon    & Improving feature descriptiveness  \\
 & Zero-shot  &   \ding{55}  &\ding{55}&    \textcircled{ }    & \textcircled{ }   &\textcircled{ } &  \textcircled{ }  &    WinCLIP, APRIL-GAN, SAA  & Utilizing external knowledge  \\ 
 \midrule
\multirow{3}{*}{\begin{tabular}[l]{@{}l@{}}\makecell[c]{Sec.~\ref{sec:3.2} \\ Data Modality}\end{tabular}} 
& 2D-aware RGB Image   &   \textcircled{ }  &\textcircled{ }    &    \ding{52}    & \ding{55}   & \textcircled{ } &  \textcircled{ }   &    MVTec AD, Eyecandies, PAD    & Optimizing imaging factors    \\
 & 3D-aware Representation   &   \textcircled{ }  &\textcircled{ }    &  \ding{55} & \ding{52}   & \textcircled{ } &  \textcircled{ }   &    MVTec 3D, Real3D, CPMF    & Learning 3D features\\
 & Multi-modality &   \textcircled{ }  &\textcircled{ }    &    \textcircled{ }   & \textcircled{ }    & \textcircled{ } &  \textcircled{ }   &    AST, BTF, M3DM, EasyNet    & Fusing multimodal features.\\ \midrule
\multirow{2}{*}{\begin{tabular}[c]{@{}l@{}}\makecell[c]{Sec.~\ref{sec:3.3} \\ Anomaly Hierarchy}\end{tabular}}   
& Structural Anomaly&   \textcircled{ }  &\textcircled{ }    &    \textcircled{ }   & \textcircled{ }   &\ding{52} &  \ding{55}   &   DSKD, GLCF, EfficientAD & Modelling local structural patterns\\
 & Semantic Anomaly&   \textcircled{ }  &\textcircled{ }    &    \textcircled{ }    & \textcircled{ }   & \ding{55} &  \ding{52}   &MVTec LOCO, ComAD, PSAD   & Modelling relationships between entities\\ 
  
 
 \bottomrule[1.5pt]
\end{tabular}
}
\end{table*}

\noindent\paragraph{Comparison to Other Surveys.} 

Recent advancements in VAD methods, particularly tailored for 2D data with structural anomalies, have been substantial, as depicted in Figure~\ref{fig:work_number}~\cite{MVTec-AD}. The prevalence of unsupervised VAD approaches is also discernible. Notably influenced by pivotal milestones such as MVTec AD~\cite{MVTec-AD}, MVTec 3D~\cite{MVTec-3D}, and MVTec LOCO~\cite{MVTec-LOCO}, sub-settings addressing the aforementioned three challenges have demonstrated promising advancements. Examinations of these emerging trends are conspicuously absent in many current VAD surveys, as underscored in Table~\ref{tab:survey}. In contrast, our survey undertakes a more comprehensive exploration, encapsulating the latest developments in VAD from diverse perspectives.


\begin{table*}[t]
    \centering
    \caption{Qualitative comparison of representative VAD methods in Sec.~\ref{sec:3.1} in terms of frame-level AUROC.}
    \label{tab:sample-number}
    \renewcommand{\arraystretch}{1.1}
    \setlength\tabcolsep{3.0pt}
    \vspace{-2mm}
    \resizebox{1.0\linewidth}{!}{

    \begin{tabular}{@{}c|cccccccccccccc@{}}
\toprule[1.5pt]
\multicolumn{1}{c}{\multirow{2}{*}{Dataset}} & \multicolumn{3}{|c}{Semi-supervised  (10-shot)} & \multicolumn{5}{c}{Unsupervised}                           & \multicolumn{3}{c}{Few-shot (8-shot)}   & \multicolumn{3}{c}{Zero-shot}                  \\ \cmidrule(l){2-4} \cmidrule(l){5-9} \cmidrule(l){10-12} \cmidrule(l){13-15} 
\multicolumn{1}{c}{}                         & \multicolumn{1}{|c}{DRA}        & PRN        & BGAD        & Draem & PatchCore & RD4AD & TFA-Net & MemKD & RegAD & {\small GraphCore} & {\small FastRecon} & WinCLIP & VAND  & {\small AnomalyCLIP}  \\ \midrule
MVTec AD                                     & 96.1          & 99.4          & 99.3           & 98.0       & 99.2         & 98.4     &      98.7 &  99.6      & 91.2     & 95.9         & 95.2         & 91.8       & 86.1      & 91.5           \\
VisA                                         & -             & -             & -              & 88.7     & 95.1         & 96.0       &    88.7  & 97.6          & -        & -            & -            & 78.1       & 78.0       & 82.1           \\ \bottomrule[1.5pt]
\end{tabular}
}
    \vspace{-5mm}
\end{table*}

\section{Taxonomy} \label{sec:3}

In this section, we review existing methods from three perspectives that correspond to the aforementioned three challenges in Table~\ref{tab:summary}.

\subsection{From the Perspective of Sample Number} \label{sec:3.1}

Given the challenge of data scarcity in practical scenarios, various VAD tasks consider varying numbers of normal and abnormal training samples. The prevalent sub-settings are shown in the top part of Table~\ref{tab:summary}.

\noindent\paragraph{Semi-supervised VAD.}
Semi-supervised VAD methods aim to utilize both normal samples and infrequent observed abnormal samples during training. However, relying solely on a restricted set of seen anomalies may induce overfitting, resulting in poor generalization capacity to novel anomalies~\cite{yao2023explicit}. To mitigate the overfitting, DRA~\cite{ding2022catching} introduced a disentanglement strategy for anomalies in open-world scenarios. This strategy classifies anomalies into three distinct categories: seen anomalies, pseudo anomalies, and latent residual anomalies. Specific detection heads are trained for individual data types, enabling them to specialize in detecting specific anomalies accordingly. Similarly, PRN~\cite{zhang2023prototypical} harnesses both seen anomalies and pseudo anomalies to explicitly capture residual features distinguishing anomalies from normal patterns. PRN employs various anomaly generation strategies, considering both seen and unseen appearance variances to create diverse pseudo anomalies. Through learning from these anomalies, PRN constructs multi-scale prototypes and generates more faithful representations for open-world anomalies rather than fixating on seen anomalies. BiaS~\cite{BiaS} also disentangles open-world anomalies into seen and unseen anomalies and utilizes two specialists for these anomalies, respectively. Then BiaS proposes a dynamic fusion strategy to fuse the prediction results of these two specialists intelligently. In contrast, BGAD~\cite{yao2023explicit} focuses on modeling the normal feature distribution through a flow model, concurrently incorporating seen anomalies to optimize the description boundary of normal features. In summary, the aforementioned methods collectively address overfitting to seen anomalies by introducing diverse pseudo anomalies or by focusing on optimizing the description boundary for normal samples through the incorporation of seen anomalies.

\noindent\paragraph{Unsupervised VAD.}
Unsupervised VAD concentrates on discerning anomalies trained exclusively on normal samples for specific categories~\cite{CDO}. The primary goal is to model the distribution of normal features, typically involving two sub-steps: feature extraction and distribution modeling. Recent advancements predominantly employ pretrained neural networks such as ResNet for feature extraction. Four main schemes for distribution modeling include memory bank, reconstruction, knowledge distillation, and flow-based methods. Memory bank-based methods, exemplified by PatchCore~\cite{Patchcore}, directly store features of training normal samples. They then utilize the nearest distance between testing samples and the stored bank to score anomalies. By selecting the most representative features in the training set, the memory bank can be small and representative, ensuring efficient and effective VAD. 
Reconstruction-based techniques, such as DFR~\cite{DFR2020}, and knowledge distillation approaches, exemplified RD4AD~\cite{RD4AD} and ViTAD~\cite{vitad}, entail the regression of extracted normal features using a secondary trainable network. In the context of reconstruction-based methods, this trainable network is referred to as autoencoders, while in knowledge distillation-based methods, it is denoted as student networks. Since the trainable network is exclusively trained with normal samples, it is expected to produce substantial regression errors for abnormal samples. 
In contrast, flow-based models~\cite{Cflowad} utilize a normalizing flow framework to automatically depict the distribution of normal features and explicitly estimate the likelihood of tested features. However, the aforementioned VAD methods often confront an issue where anomaly scores for abnormalities can unexpectedly be low due to imprecise boundary descriptions, attributed to the generalization ability of employed neural networks, termed as over-generalization in CDO~\cite{CDO}. To mitigate over-generalization, some methods like Draem~\cite{Draem}  MRKD~\cite{jiang_masked_2023}, and DAF~\cite{2023CaiDis} introduce synthetic anomalies. Consequently, models are not only tasked with regressing normal feature distributions but also with generating substantial regression errors for synthetic anomalies. 
MemKD~\cite{gu2023remembering} addresses the over-generalization problem by explicitly storing a memory bank, ensuring that outputs exclusively represent normal features.  Similarly, TFA-Net~\cite{TFA-Net} proposes to restore normal features explicitly guided by normal templates. These approaches result in significant regression errors when faced with abnormal inputs.

\noindent\paragraph{Few-shot VAD.}
Few-shot VAD concentrates on training the model with a limited amount of normal data~\cite{regad}. However, these few normal samples may not adequately represent the entire normal sample set. Consequently, the model must establish a description boundary by learning from these seen normal samples, intending to simultaneously describe distributions of unseen normal samples while excluding distributions of abnormal samples. This presents a notable challenge. To address this challenge, few-shot VAD methods primarily emphasize enhancing feature descriptiveness, aiming to make the available few-shot samples a more representative subset. RegAD~\cite{regad} employs registration-based proxy tasks for representation learning, aligning samples of the same category through geometrical transformations, thereby enhancing feature descriptiveness. Another method, GraphCore~\cite{Graphcore}, utilizes a vision isometric invariant graph neural network to extract rotation-invariant structural features, particularly beneficial for categories with geometrical transformations. Additionally, FastRecon~\cite{fang2023fastrecon} proposes utilizing a few normal samples as a reference to reconstruct their normal versions, achieving anomaly detection through sample alignment. In comparison to RegAD, FastRecon also aims to align few-shot normal samples and testing samples, while accommodating more complex geometrical transformations. In summary, prevailing few-shot VAD methods commonly rely on sample alignment to augment feature descriptiveness, aiming to enhance the representativeness of seen normal samples for the entire normal sample set.

\noindent\paragraph{Zero-shot VAD.}
Zero-shot VAD aims to develop a unified model for detecting anomalies across diverse domains without relying on reference normal samples~\cite{WinClip}. While holding significant potential for versatility, this task presents a challenge due to the absence of specific prior information related to the target domains. Existing zero-shot VAD methods tackle this challenge by integrating external knowledge to enhance anomaly detection capabilities. WinCLIP~\cite{WinClip}, a pioneering zero-shot VAD method, leverages a pretrained visual-language model (VLM) CLIP~\cite{clip}. Through the integration of CLIP, WinCLIP achieves zero-shot VAD by computing the similarity between image patches and normal/abnormal textual captions. Since CLIP has acquired implicit knowledge for distinguishing normality from anomalies through training on extensive datasets with visual-text pairs, the calculated similarities can serve as effective anomaly scores. 
APRIL-GAN~\cite{VAND}, succeeding WinCLIP, tackles the domain gap between CLIP and targeted VAD data. APRIL-GAN proposes adapting CLIP to VAD by training it with annotated auxiliary VAD data, thereby enhancing its suitability for VAD applications. Building on this adaptation scheme, AnomalyCLIP~\cite{AnomalyCLIP} introduces the concept of learning object-agnostic text prompts to overcome the limitations associated with manually designed prompts. Additionally, SAA~\cite{SAA} introduces an ensemble scheme that combines various off-the-shelf VLMs for VAD, providing a means to integrate human expertise into VAD systems. In summary, these zero-shot VAD methods leverage external knowledge, often from off-the-shelf VLMs like CLIP, to facilitate the detection of anomalies in arbitrary categories.

\noindent\paragraph{Discussion.}
This section evaluates the discussed methods from the perspective of sample numbers. As we transition from semi-supervised to zero-shot VAD, the available training samples for specific categories gradually diminish, resulting in a decline in VAD performance, as illustrated in Table~\ref{tab:sample-number}. While VAD performance reaches saturation with ample samples, the effectiveness of few-shot and zero-shot VAD remains suboptimal.
Nevertheless, the shared objective across these sub-settings is to learn from available data and develop a VAD model capable of generalizing to unseen normal samples while detecting abnormal samples. Given the variability in accessible samples within individual sub-settings, their focal points may differ, leading to distinct key motivations. Future efforts could be directed toward establishing a unified VAD framework for all these sub-settings, thus efficiently utilizing available data and consolidating efforts to identify optimal methods for practical applications.

\subsection{From the Perspective of Data Modality} \label{sec:3.2}
In this section, we categorize VAD from the perspective of data modality, as shown in the middle part of Table~\ref{tab:summary}.

\noindent\paragraph{2D-aware RGB Image.} 
RGB images play a crucial role in VAD. Datasets like MVTec AD~\cite{MVTec-AD} and VisA~\cite{VisA} have curated extensive datasets containing both normal and abnormal images, significantly driving advancements for VAD on RGB images, as Sec.~\ref{sec:3.1} elaborates. Despite the commendable performance on this benchmark, these datasets typically assume an ideal imaging environment with perfectly aligned objects and optimal illumination. These critical imaging factors are gradually taken into consideration. Illumination, a critical aspect of imaging systems, significantly influences imaging quality~\cite{Eyecandies}. Adequate illumination conditions can enhance the visibility of anomalies. Eyecandies~\cite{Eyecandies} accounts for different illumination conditions. Specifically, four lights in various positions are employed to capture multi-illumination images in Eyecandies. A simple autoencoder is then applied to these images, along with depth and normal images~\cite{Eyecandies}. Considering the pose of objects, most categories in MVTec AD assume perfectly aligned objects, which may not reflect real-world scenarios where objects can be in any pose. PAD~\cite{PAD} introduces a multi-pose VAD dataset and formulates a pose-agnostic VAD task. In summary, there have already been plenty of methods proposed for RGB images, especially for datasets like MVTec AD that are equipped with ideal imaging environments. Recent research has focused on exploring VAD for RGB images in practical imaging environments, considering factors such as unideal illumination and diverse poses.

\noindent\paragraph{3D-aware Representation.} 
Geometrical information, often represented as 3D data like point clouds, serves as a direct manifestation of the size and shape of visual entities. Two notable datasets designed for point cloud VAD are MVTec 3D~\cite{MVTec-3D} and Real3D~\cite{Real3D}. These datasets encompass high-resolution point clouds, facilitating the identification of subtle geometrical deviations. Similar to image VAD, point cloud VAD methods can be broadly categorized into two sub-steps~\cite{CPMF}: feature extraction and distribution modeling. In contrast to the image domain, where numerous effective off-the-shelf pretrained models can serve as feature extractors, pretrained point cloud neural networks lack robustness~\cite{3DST}. Therefore, a self-supervised learning scheme introduced in~\cite{3DST} aims to construct a more resilient feature extractor for point cloud VAD. Subsequently, a knowledge distillation-based approach is employed for distribution modeling. On the other hand, CPMF~\cite{CPMF} transforms point clouds into multi-view depth images, enabling the use of established pretrained image models for point cloud feature extraction. CPMF then incorporates PatchCore~\cite{Patchcore} for distribution modeling. In summary, unlike the abundance of off-the-shelf models for RGB images, there are limited robust models for other modalities. Existing VAD methods that consider the 3D modality usually place special emphasis on learning descriptive features.

\noindent\paragraph{Multi-modality.} 
In specific scenarios, employing multi-modality data enhances the comprehensiveness of VAD, such as the coexistence of 3D and RGB modalities. Some methods are explicitly designed to improve the fusion of representations from these modalities. For example, BTF~\cite{BTF} straightforwardly concatenates 3D and RGB representations, utilizing them as inputs for PatchCore~\cite{Patchcore}. Building upon BTF, M3DM~\cite{m3dm} incorporates contrast learning on the two modalities, fostering enhanced synergy between them. Similarly, Shape-Guided~\cite{Shape-Guided} integrates the two representations guided by shape features. However, these methods heavily rely on pretrained networks, potentially lacking robustness, especially in the context of point cloud networks~\cite{BTF}. In contrast, AST~\cite{AST} chooses to learn point cloud representation directly from raw data by training an asymmetric teacher-student pair. This pair can process both RGB and 3D data, leading to better integration between the two modalities. Additionally, EasyNet~\cite{EasyNet} and 3DSR~\cite{3DSR} generate synthetic abnormal RGB and point cloud data to train a robust feature extractor for both modalities. In essence, these methods for multi-modality typically concentrate on enhancing the learning and fusion of representations across multiple modalities.


\noindent\paragraph{Discussion.}
Data modalities for VAD may exhibit variation across targeted scenarios. Among these modalities, RGB and 3D stand out as dominant and extensively explored. In the progression of VAD comprehension, various factors, such as illumination~\cite{Eyecandies} and pose~\cite{PAD}, are considered in conjunction with RGB and 3D modalities. Because of off-the-shelf pretrained models for RGB data, we have witnessed significant progress in 2D VAD. However, due to the scarcity of pretrained models for other modalities, VAD on other modalities remains less promising, as depicted in Table~\ref{tab:modality}. Table~\ref{tab:modality} further illustrates that employing multiple modalities enables a more comprehensive acquisition of real-world information, consequently leading to improved VAD performance. Future endeavors might be directed toward developing a unified VAD architecture capable of processing all these modalities, thereby enhancing multi-modality feature learning and fusion.


\begin{table}[t]
    \centering
    \caption{Qualitative comparison of representative VAD methods in Sec.~\ref{sec:3.2} in terms of frame-level AUROC.}
    \label{tab:modality}
    \renewcommand{\arraystretch}{1.1}
    \setlength\tabcolsep{3.0pt}
    \vspace{-2mm}
    \resizebox{1.0\linewidth}{!}{

            \begin{tabular}{@{}c|ccccccccc@{}}
                \toprule[1.5pt]
                \multicolumn{1}{c}{\multirow{2}{*}{Dataset}} & \multicolumn{3}{|c}{2D-aware RGB Image} & \multicolumn{3}{c}{3D-aware   Representation}  & \multicolumn{3}{c}{Multi-modality} \\ \cmidrule(l){2-4} \cmidrule(l){5-7} \cmidrule(l){8-10} 
                \multicolumn{1}{c}{}                         & \multicolumn{1}{|c}{AST}      & CDO     & M3DM      & BTF & CPMF & M3DM  & AST    & BTF    & M3DM    \\ \midrule
                MVTec 3D                                     & 88.0          & 93.8       & 85.0             & 78.2   & 95.2    & 87.4     & 93.7      & 86.5      & 94.5       \\
                Real3D                                       & -           & -          & -               & 59.3   & 62.5    & 59.4     & -         & -         & -          \\ \bottomrule[1.5pt]
                \end{tabular}
    }
    \vspace{-5mm}
\end{table}


\subsection{From the Perspective of Anomaly Hierarchy} \label{sec:3.3}
Based on the hierarchy of anomalies, the existing VAD methods can be categorized into: structural and semantic anomaly detection, as summarized in the bottom part of Table~\ref{tab:summary}.

\noindent\paragraph{Structural Anomaly.} 
Structural anomalies refer to local structural deviations like scratches, distorted shapes, etc. In recent years, there has been a notable increase in the development of VAD methods specifically tailored for addressing structural anomalies. These methods aim to learn fine-grained features that comprehensively describe local structural patterns within visual entities. Almost all the methods mentioned~\cite{CPMF,Patchcore} above focus on structural anomaly detection. A recent noteworthy contribution in this domain is EfficientAD~\cite{Batzner2023EfficientADAV}, which proposes a lightweight encoder generated through knowledge distillation. This lightweight encoder restricts the receptive field to a relatively small region, enhancing the modeling of local structures and facilitating both efficiency and effective anomaly detection.

\begin{table}[t]
    \centering
    \caption{Qualitative comparison of representative VAD methods in Sec.~\ref{sec:3.3} in terms of frame-level AUROC. }
    \label{tab:hierachy}
    \renewcommand{\arraystretch}{1.1}
    \setlength\tabcolsep{3.0pt}
    \vspace{-2mm}
    \resizebox{1.0\linewidth}{!}{

        \begin{tabular}{@{}c|cccccccc@{}}
            \toprule[1.5pt]
            \multirow{2}{*}{Dataset} & \multicolumn{4}{c}{Structural Anomaly}                  & \multicolumn{4}{c}{Semantic Anomaly}                    \\ \cmidrule(l){2-5} \cmidrule(l){6-9}  
                                     & GLCF & {\small EfficientAD} & ComAD  & PSAD & GLCF & {\small EfficientAD} & ComAD  & PSAD \\ \midrule
            MVTec LOCO               & 83.8 & 94.7           & 90.9         & 91.6     & 82.4   & 86.8           & 89.4         & 98.1   \\ \bottomrule[1.5pt]
            \end{tabular}
    }
    \vspace{-5mm}
\end{table}


\noindent\paragraph{Semantic Anomaly.} 

In contrast to structural anomalies occurring within individual visual entities, semantic anomalies manifest in the relations between multiple entities within a frame. To advance research in this area, datasets like MVTec LOCO~\cite{MVTec-LOCO} have been introduced. These datasets typically involve multiple co-occurring visual entities, where their relations may exhibit abnormalities. Various approaches have been proposed to address semantic anomalies. One line of research posits that global information can implicitly capture the relations between entities. For instance, GCAD~\cite{MVTec-LOCO} operates on both local-local consistencies and global-global consistencies through two student-teacher pairs. Specifically, one student-teacher pair with a small receptive field compares local structural features in a regression-based manner to identify structural anomalies. Another student-teacher pair, capable of extracting global context for the given frame, similarly compares global semantic features to identify semantic anomalies. Subsequent methods, such as DSKD~\cite{zhang_contextual_2024}, GLCF~\cite{Yao2023LGC}, and EfficientAD~\cite{Batzner2023EfficientADAV}, build upon GCAD and also model normal relations between entities through local-local and global-global consistencies. These methods employ various strategies to enhance the understanding of the normal global context, including contextual affinity distillation~\cite{zhang_contextual_2024} and local-global alignment~\cite{Yao2023LGC}. The bottleneck-like architectures in these methods play a crucial role in learning global context. 
Alternatively, some methods explicitly focus on modeling the relations between entities. ComAD~\cite{ComAD} and PSAD~\cite{kim_few_2024} initially detect individual entities within frames. In the absence of accessible entity-level labels, they typically use cluster-based methods to group similar data points into clusters, thereby detecting entities. Subsequently, they model the relations between entities through strategies like histogram analysis. This approach allows for a more precise description of the relations between entities. In summary, existing methods for semantic anomaly build an understanding of relations between entities either implicitly by learning global context or explicitly by extracting relations between entities.

\noindent\paragraph{Discussion.}

Anomalies can manifest at various hierarchical levels. While early efforts primarily focused on modeling local structural contexts for structural VAD, there has been a gradual shift in the popularity of semantic VAD. Semantic VAD considers abnormal relations among visual entities. Existing methods in semantic VAD predominantly emphasize modeling relations between entities. Some approaches aim to implicitly capture these relations through global context, while others explicitly analyze individual entities and their relations. However, current methods may encounter challenges in understanding normal relations as effectively as humans, particularly for intricate relations involving entity numbers, locations, etc. The implicit learning strategy may fall short in precisely identifying abnormal relations through global context. In contrast, the explicit strategy may better delineate the relations and yield more precise detection results. As depicted in Table~\ref{tab:hierachy}, explicit methods such as ComAD and PSAD outperform implicit methods like GLCF and EfficientAD in addressing semantic anomalies. Nevertheless, the methods for entity extraction and relation modeling still leave ample room for improvement.



\subsection{Other Perspectives} 

In addition to the previously discussed major perspectives, there are other settings that warrant exploration. For example, noisy VAD posits that practical application labels for training samples may contain errors, potentially impacting the efficacy of conventional VAD methods. SoftPatch~\cite{SoftPatch} suggests denoising data at the patch level by generating soft outlier scores for patches, thereby excluding patches with high outlier scores, \textit{i.e.}, noisy data, for training. Continual VAD~\cite{UCAD}, on the other hand, endeavors to enhance VAD models with gradually accessible novel data. Directly updating VAD models with this data may result in catastrophic forgetting and impose a substantial computational burden. Hence, UCAD~\cite{UCAD} proposes empowering VAD models with continual learning capacity by establishing a key-prompt-knowledge memory space. Uniformed VAD~\cite{DiAD} has also gained recent popularity, aiming to construct a unified VAD model for diverse categories. Unlike zero-shot VAD, which operates without data from targeted categories, uniformed VAD focuses on effectively utilizing samples from specific categories. DiAD~\cite{DiAD}, as a representative uniformed VAD method, suggests using diffusion models to restore normal references for testing samples. The disparities between testing samples and the restored normal references in the feature space are then utilized to score anomalies. Considering the scarcity of VAD data, some methods~\cite{dai_generating_2024} also endeavor to generate VAD data. In summary, there are numerous variants to consider for practical VAD systems.

\section{Future Directions}\label{sec:4}
This survey provides a comprehensive illustration of the evolution of VAD methods from diverse perspectives, highlighting the challenges faced by existing methods. The subsequent discussion will delve into forthcoming trends across the aforementioned varied perspectives.

\subsection{Towards Generic VAD}

Current literature has focused on developing VAD methods under varying sample numbers, attributed to the diverse accessible samples in different scenarios. Future efforts could be directed toward constructing a generic VAD framework capable of accommodating different sample numbers.

\noindent\paragraph{Foundation Model for VAD.} Recently, foundation models like GPT4-V(ision)~\cite{gpt4v} and SAM~\cite{SAM} have demonstrated outstanding generalization abilities, showcasing scalable performance with different sample numbers. These foundation models also present some efficacy for VAD~\cite{VAND,SAA,gpt-4v-ad}. Advanced techniques, such as prompting learning~\cite{MaPLe} could further enhance the performance of foundation models for VAD. Additionally, training a foundation model specifically tailored for VAD may yield more promising VAD performance. Various pretraining schemes, like contrastive learning~\cite{clip}, and sequential modeling~\cite{bai2023sequential} can be explored. The in-context learning capacity of these foundational VAD models should be taken into consideration, enabling adaptations with limited data without additional training.

\noindent\paragraph{Scalable Data for VAD.} The availability of large-scale data is crucial for constructing foundation models for VAD. Practical improvements in visual data collection are warranted. On the other hand, anomaly generation can further contribute to scalable data. While some works like DFMGAN~\cite{Duan2023DFMGAN}, and AnomalyDiffusion~\cite{hu2023anomalydiffusion} have conducted preliminary investigations into anomaly generation, their current generalization ability is insufficient. In the field of image generation, methods like ControlNet~\cite{ControlNet} have demonstrated robust generalization abilities and fine-grained control over the generation process. Future endeavors should focus on developing data generation methods capable of faithfully producing anomalies across a wide range of categories, contributing to the establishment of scalable data for VAD.

\subsection{Towards Multimodal VAD}
Multimodal data can comprehensively reflect the information of visual entities, leading to enhanced VAD performance. Looking ahead, more attention can be directed towards joint imaging parameter optimization and multimodal learning.

\noindent\paragraph{Imaging Parameter Optimization for VAD.}
Conventional public VAD datasets~\cite{MVTec-AD,MVTec-3D} commonly assume ideal imaging conditions. However, challenges emerge in complex practical scenarios where given imaging parameters are unideal. While Eyecandies~\cite{Eyecandies} and PAD~\cite{PAD} have explored the impact of imaging parameters, a comprehensive analysis on optimizing the imaging process is lacking. Imaging parameter optimization strives to automatically optimize the imaging parameters, encompassing aspects like auto-exposure, auto-focus, etc. This technique opens new avenues for acquiring higher-quality data, facilitating improved visualization, and enabling easier detection of anomalies.

\noindent\paragraph{Multimodal Learning for VAD.}
The importance of effective representations in VAD is crucial~\cite{vitad}, particularly in multimodal data~\cite{CPMF}. Achieving effective fusion between these modalities is essential for reliable VAD. In contrast to multimodal fusion in other fields~\cite{MultimodalLearning}, existing integration methods in VAD appear relatively simplistic, like feature concatenation~\cite{m3dm,BTF}. Considering multiple modalities for VAD in real-world applications, it is promising to construct a unified architecture capable of multimodal data. In this context, VAD data from different scenarios and modalities can be utilized together to build a unified VAD model, enhancing feature learning and fusion.

\subsection{Towards Holistic VAD}

While structural VAD has shown promising performance~\cite{CDO}, the capability to detect semantic anomalies is crucial for practical VAD systems. From a broader perspective, VAD systems must not only identify anomalies but also establish connections to downstream processes, leading to better overall performance.

\noindent\paragraph{Understanding of Relations between Entities.}
Semantic anomalies, distinct from structural anomalies requiring local structural representations, demand VAD models to genuinely comprehend the relations between entities. While current VAD methods for semantic anomalies~\cite{Yao2023LGC,MVTec-LOCO} demonstrate reasonable semantic VAD performance, they still fall short of truly understanding normal relations between entities. Foundation models, such as GPT-4V~\cite{gpt4v}, exhibit logical reasoning capacity, showcasing a genuine understanding of normal relations between entities~\cite{GenericAd}. Therefore, incorporating such foundation models for semantic anomaly detection appears promising. Multi-modal inputs and multi-round conversations~\cite{Dawn-of-LMMs} may further enhance the understanding of relations between entities. Additionally, it is straightforward and promising to detect visual entities first and then identify their relations, like ComAD~\cite{ComAD}. API-based modeling like ViperGPT~\cite{suris2023vipergpt} may enhance this scheme.

\noindent\paragraph{Connecting VAD with Downstream Tasks.}
VAD plays a pivotal role in interconnected systems, particularly in quality inspection pipelines. However, current research often focuses solely on enhancing the isolated perception step, neglecting downstream integration and impact. To optimize effectiveness, VAD must be seamlessly integrated into the broader system workflow. A comprehensive understanding is required of how VAD interacts with and relies on other components, including potential feedback loops. Recent work has initiated exploration in this direction by incorporating VAD outcomes into objectives such as robotic navigation~\cite{wellhausen2020safe} and manufacturing processes~\cite{Singh2023TowardCA}. Moving forward, greater emphasis should be placed on end-to-end system optimization and unified representation learning achieved through the tight integration of VAD and downstream tasks.

\section{Conclusion}\label{sec:5}
In this survey, we have delved into recent advancements in Visual Anomaly Detection (VAD). Initially, we underscored three challenges: 1) scarcity of training data, 2)  diversity of visual modalities, and 3)  complexity of hierarchical anomalies. Subsequently, we furnished background information on VAD and delved into key concepts. Following that, we conducted a comprehensive review of existing VAD methods, with a focus on sample numbers, data modalities, and anomaly hierarchies. Finally, we pinpointed promising directions for future research: generic, multimodal, and holistic VAD. We anticipate that tackling these challenges and pursuing these research directions will propel VAD towards more robust deployment in real-world applications.


\clearpage

\clearpage

\appendix





\bibliographystyle{named}
\begin{small}

\end{small}

\end{document}